\newif\iftwocol\twocoltrue      

\documentclass[twoside\iftwocol,twocolumn\else,12pt\fi]{article}
\usepackage{latexsym}
\def\,{\mskip 3mu} \def\>{\mskip 4mu plus 2mu minus 4mu} \def\;{\mskip 5mu plus 5mu} \def\!{\mskip-3mu}
\def\dispmuskip{\thinmuskip= 3mu plus 0mu minus 2mu \medmuskip=  4mu plus 2mu minus 2mu \thickmuskip=5mu plus 5mu minus 2mu}
\def\textmuskip{\thinmuskip= 0mu                    \medmuskip=  1mu plus 1mu minus 1mu \thickmuskip=2mu plus 3mu minus 1mu}
\textmuskip
\def\beq{\dispmuskip\begin{equation}}    \def\eeq{\end{equation}\textmuskip}
\def\beqn{\dispmuskip\begin{displaymath}}\def\eeqn{\end{displaymath}\textmuskip}
\def\bqa{\dispmuskip\begin{eqnarray}}    \def\eqa{\end{eqnarray}\textmuskip}
\def\bqan{\dispmuskip\begin{eqnarray*}}  \def\eqan{\end{eqnarray*}\textmuskip}

\renewenvironment{abstract}{\begin{quote}{\noindent\bf Abstract.}}{\end{quote}}
\newenvironment{keywords}{\begin{quote}{\noindent\bf Keywords.}}{\end{quote}}

\newtheorem{theorem}{Theorem}
\newtheorem{corollary}[theorem]{Corollary}
\newtheorem{lemma}[theorem]{Lemma}
\newtheorem{definition}[theorem]{Definition}

\def\ftheorem#1#2#3{\begin{theorem}[#2]\label{#1} \it #3 \end{theorem} }
\def\fcorollary#1#2#3{\begin{corollary}[#2]\label{#1} \it #3 \end{corollary} }

\def\subsection#1{\vspace{1ex}\noindent{\bf #1.}}
\def\toinfty#1{\stackrel{#1\to\infty}{\longrightarrow}}
\def\nq{\hspace{-1em}}
\def\qed{\hspace*{\fill}$\Box\quad$}
\def\odt{{\textstyle{1\over 2}}}
\def\odf{{\textstyle{1\over 4}}}
\def\odA{{\textstyle{1\over A}}}
\def\odn{{\textstyle{1\over n}}}
\def\eps{\varepsilon}
\def\vec#1{{\bf #1}}
\def\M{{\cal M}}
\def\X{{\cal X}}
\def\Y{{\cal Y}}

\def\E{{\bf E}}
\def\P{{\bf P}}
\def\Set#1{\if#1Q{I\!\!\!#1}\else\if#1Z{Z\!\!\!Z}\else{I\!\!#1}\fi\fi}
\def\qmbox#1{{\quad\mbox{#1}\quad}}
\def\sumprime{\mathop{{\sum\nolimits'}}}
\def\tolim{\mathop{\longrightarrow}\limits}

\begin{document}

\title{\vspace*{-2cm}
   {\normalsize Technical Report IDSIA-09-01 \hfill 15 August 2001 -- 16 January 2003\\[1mm]}
  \huge\sc\hrule height1pt \vskip 2mm
     Convergence and Loss Bounds for \\ Bayesian Sequence Prediction
     \vskip 2mm \hrule height1pt}

\author{Marcus Hutter \\[1ex]
\small IDSIA, Galleria 2, CH-6928 Manno-Lugano, Switzerland%
\thanks{This work was supported by
SNF grant 2000-61847.00 to J\"urgen Schmidhuber.} \\
\small marcus@idsia.ch, http://www.idsia.ch/$^{_{_\sim}}\!$marcus
}

\date{}

\maketitle              

\begin{abstract}
The probability of observing $x_t$ at time $t$, given past
observations $x_1...x_{t-1}$ can be computed with Bayes' rule if
the true generating distribution $\mu$ of the sequences
$x_1x_2x_3...$ is known. If $\mu$ is unknown, but known to belong
to a class $\M$ one can base ones prediction on the Bayes mix
$\xi$ defined as a weighted sum of distributions
$\nu\in\M$.
Various convergence results of the mixture posterior $\xi_t$ to
the true posterior $\mu_t$ are presented. In particular a new
(elementary) derivation of the convergence $\xi_t/\mu_t\to 1$ is
provided, which additionally gives the rate of convergence.
A general sequence predictor is allowed to choose an action $y_t$
based on $x_1...x_{t-1}$ and receives loss $\ell_{x_t y_t}$ if $x_t$
is the next symbol of the sequence. No assumptions are made on the
structure of $\ell$ (apart from being bounded) and $\M$.
The Bayes-optimal prediction scheme $\Lambda_\xi$ based on mixture
$\xi$ and the Bayes-optimal informed prediction scheme
$\Lambda_\mu$ are defined and the total loss $L_\xi$ of
$\Lambda_\xi$ is bounded in terms of the total loss $L_\mu$ of
$\Lambda_\mu$. It is shown that $L_\xi$ is bounded for bounded
$L_\mu$ and $L_\xi/L_\mu\to 1$ for $L_\mu\to \infty$. Convergence
of the instantaneous losses are also proven.
\end{abstract}


\section{Introduction}\label{secInt}

\subsection{Setup}
We consider inductive inference problems in the following form:
Given a string $x_1 x_2 ... x_{t-1}$, we want to predict its
continuation $x_t$. We assume that the strings which have to
be continued are drawn from a probability distribution $\mu$. The
maximal prior information a prediction algorithm can possess is
the exact knowledge of $\mu$, but in many cases the true
generating distribution is not known.
In order to overcome this problem a mixture distribution $\xi$ is
defined as a $w_\nu$ weighted sum over distributions $\nu\in\cal
M$, where $\cal M$ is any discrete (hypothesis) set including
$\mu$. We assume that $\M$ is known and contains the true
distribution, i.e. $\mu\in\M$.
Since the posterior $\xi_t$ can be shown to converge rapidly to
the true posterior $\mu_t$, making decisions based on $\xi$ is
often nearly as good as the infeasible optimal decision based on
the unknown $\mu$  \cite{Feder:98}. In this work we compare the
expected loss of predictors based on mixture $\xi$ to the expected
loss of informed predictors based on $\mu$.

\subsection{Contents}
Section \ref{secSetup} introduces concepts and notation needed
later, including strings, probability distributions, mixture
distributions, expectations, and various types of convergence and
distance measures.
Section \ref{secConv} summarizes various convergence results of
the mixture distribution $\xi$ to the true distribution $\mu$. We
provide a new (elementary) derivation of the posterior convergence
in ratio, which is not based on Martingales, but on the Hellinger
distance, and compare it to related known results
\cite{Doob:53,Li:97,Vovk:87,Vitanyi:00}.
Section \ref{secLoss} introduces the decision theoretic setup,
where an action/prediction $y_t$ results in a loss $\ell_{x_t
y_t}$ if $x_t$ is the next symbol of the sequence. Improving upon
previous results in \cite{Feder:98,Hutter:01alpha,Hutter:01loss},
the expected total (or cumulative) loss $L_\xi$ made by the
Bayes-optimal prediction scheme based on mixture $\xi$ minus the
expected total loss $L_\mu$ of the optimal informed prediction
scheme based on $\mu$ is bounded by $O(\sqrt{L_\mu})$. Some
popular loss functions, including the absolute, square,
logarithmic, Hellinger, and error loss are discussed.
A Proof of the loss bound is given in Section \ref{secLProof}.
Convergence of the instantaneous losses are briefly studied in
Section \ref{secILoss}.
Section \ref{secConc} recapitulates the assumptions made in this
work and possible relaxations, mentions some optimality properties
of $\xi$ proven in \cite{Hutter:01op}, and provides an outlook to
future work.

\section{Preliminaries}\label{secSetup}

\subsection{Strings and Probability Distributions}
We denote strings over a finite alphabet $\X$ by $x_1x_2...x_n$
with $x_t\in\X$. We abbreviate $x_{n:m}:=x_nx_{n+1}...x_{m-1}x_m$
and $x_{<n}:=x_1... x_{n-1}$. We use Greek letters for probability
distributions/measures, especially $\rho$ for arbitrary ones,
$\mu\in\M$ for the true (generating) one, $\nu\in\M$ for arbitrary
ones in $\M$, and $\xi$ for the mixture (\ref{xidefsp}). Let
$\rho(x_{1:t})$ be the probability that an (infinite) sequence
starts with $x_1...x_t$. The conditional $\rho$ probability that a
given string $x_1...x_{t-1}$ is continued by $x_t$ is
$\rho_t:=\rho(x_t|x_{<t})=\rho(x_{1:t})/\rho(x_{<t})$.
The considered prediction schemes will be based on these posteriors.

\subsection{Mixture distributions}
Let $\M := \{\mu_1,\mu_2,...\}$ be a finite or countable set of
candidate probability distributions on strings. We
define a weighted average on $\M$
\beq\label{xidefsp}
  \xi(x_{1:n}) \;:=\;
  \sum_{\nu\in\M}w_\nu\!\cdot\!\nu(x_{1:n}),\quad
  \sum_{\nu\in\M}w_\nu=1,\quad w_\nu>0.
\eeq
$\xi$ is called a Bayes-mixture. The weights $w_\nu$ may be
interpreted as the prior belief in environment $\nu\in\M$.
The most interesting property the mixture distribution $\xi$
is that it multiplicatively dominates all
distributions in $\M$:
\beq\label{unixi}
  \xi(x_{1:n}) \;\geq\;
  w_\nu\!\cdot\!\nu(x_{1:n}) \quad\mbox{for all}\quad
  \nu\in\M.
\eeq
In the following, we assume that $\M$ is known and contains the
true distribution, i.e. $\mu\in\M$.
If $\M$ is chosen sufficiently large, then $\mu\in\M$ is not a
serious constraint. Generic classes, especially where $\M$
contains {\em all} (semi)computable probability distributions are
discussed in
\cite{Solomonoff:78,Li:97,Hutter:01alpha,Hutter:01op}.
Generalizations to the case where $\M$ does not contain $\mu$ are
briefly discussed in \cite{Hutter:01op} and more intensively in
a related context in \cite{Gruenwald:98}.

\subsection{Expectations and convergence measures}
We use $\E[..]$ to denote expectations w.r.t.\ the ``true'' distribution
$\mu$ and abbreviate $\E_t[..]:=\E[..|x_{<t}]$.
If $[..]$ depends on $x_{1:t}$ only, i.e.\ is independent of
$x_{t+1:\infty}$, we have
\beqn
\E[..]:=\nq\sumprime_{x_{1:t}\in\X^t}\!\!\!\mu(x_{1:t})[..]
\qmbox{and}
\E_t[..]:=\sumprime_{x_t\in\X}\mu(x_t|x_{<t})[..],
\eeqn
where $\sumprime$ sums over all $x_t$ or $x_{1:t}$ for which
$\mu(x_{1:t})\neq 0$. Similarly we use $\P[..]$ to denote the
$\mu$ probability of event $[..]$. We need the following kinds of
convergence of a random sequence $z_1,z_2,...$ to (a random
variable) $z_*$:
\bqan
\mbox{with probability 1}
& \mbox{(w.p.1)} & \P[z_t\!\toinfty{t}\!z_*]=1 \\[-0.5ex]
\mbox{in probability}
& \mbox{(i.p.)} & \forall\eps\!:\!\P[|z_t-z_*|\geq\eps]\toinfty{t} 0 \\
\mbox{in mean sum}
& \mbox{(i.m.s.)} & \textstyle\sum_{t=1}^\infty\E[(z_t-z_*)^2]<\infty \\[-0.5ex]
\mbox{in the mean}
& \mbox{(i.m.)} & \E[(z_t-z_*)^2]\toinfty{t} 0
\eqan
Convergence in one sense may imply convergence in another sense.
The following implications are valid, strict, and complete:
\beqn
  i.m.s.
  \;{\textstyle{\nearrow\atop\searrow}}
  {w.p.1\atop i.m.}
  \!{\textstyle{\searrow\atop\nearrow}}\;
  i.p.
\eeqn
Convergence i.m.s.\ is very strong: it
provides a rate of convergence in the sense that the expected
number of times $t$ in which $z_t$ deviates more than $\eps$ from
$z_*$ is finite and bounded by
$\sum_{t=1}^\infty\E[(z_t-z_*)^2]/\eps^2$.

\subsection{Distance Measures}
We need several distance measures between probability distributions
$y_i\geq 0$, $z_i\geq 0$, $\sum_i y_i=\sum_i z_i=1$, $i=\{1,...,N\}$,
namely the
\bqa\label{vecdist}
  \mbox{absolute distance:}
     & &\nq\textstyle a=\sum_i|y_i-z_i| \\[-0.5ex]\nonumber
  \mbox{square or Euclidian distance:}
     & &\nq\textstyle s=\sum_i(y_i-z_i)^2 \\[-0.5ex]\nonumber
  \mbox{Hellinger distance:}
     & &\nq\textstyle h=\sum_i(\sqrt{y_i}-\sqrt{z_i})^2 \\[-0.5ex]\nonumber
  \mbox{relative entropy or KL divergence:}
     & &\nq\textstyle d=\sum_i y_i\ln{y_i\over z_i} \\[-0.5ex]\nonumber
  \mbox{absolute divergence:}
     & &\nq\textstyle b=\sum_i y_i|\ln{y_i\over z_i}|
\eqa
All bounds we prove in this work heavily rely on the following
inequalities:
\beq\label{entro2ineqn}\label{entroaineqn}\label{entroineq2}\label{lemEIneq}
  s \;\leq\; d,\qquad
  h \;\leq\; d,\qquad
b-d \;\leq\; a \;\leq\; \sqrt{2d}.
\eeq
See \cite{Hutter:01alpha}, \cite[Lem.12.6.1]{Cover:91}, and
\cite[p178]{Borovkov:98} for proofs of $s\leq d$,
$a\leq\sqrt{2d}$, and $h\leq d$, respectively. $b-d\leq a$ is
elementary and follows from $-\ln x\leq{1\over x}-1$.
Inequality $s\leq d$ is a generalization of the binary $N=2$ case
used in \cite{Solomonoff:78,Hutter:99errbnd,Li:97}. If we insert
\beq\label{xydef}
  \X=\{1,...,N\},\quad
  N=|\X|, \quad
  i=x_t,
\eeq\vspace{-3ex}
\beq\label{defmuxit}
  y_i=\mu_t:=\mu(x_t|x_{<t}), \quad
  z_i=\xi_t:=\xi(x_t|x_{<t})
\eeq
into (\ref{vecdist}) we get various {\em instantaneous distances}
(at time $t$) between $\mu$ and $\xi$. If we take the expectation
(over $x_{<t}$) and sum over $t=1..n$, ($\sum_{t=1}^n\E[...]$)
we get various {\em total distances} between $\mu$ and $\xi$:
\beq\label{distdD}
\begin{array}{rclrcl}
\label{atmuxi}\label{ANmuxi}
  a_t(x_{<t}) \;\nq&:=&\nq\; \sum_{x_t}
  |\mu_t-\xi_t|,
  & A_n \;\nq&:=&\nq\; \sum_{t=1}^n\E[a_t]
  \\[1ex]
\label{stmuxi}\label{SNmuxi}\label{eukdistxi1}
  s_t(x_{<t}) \;\nq&:=&\nq\; \textstyle \sum_{x_t}
  (\mu_t-\xi_t)^2,
  & S_n \;\nq&:=&\nq\; \sum_{t=1}^n\E[s_t]
  \\[1ex]
\label{htmuxi}\label{HNmuxi}
  h_t(x_{<t}) \;\nq&:=&\nq\; \textstyle \sum_{x_t}
  (\sqrt{\mu_t}\!-\!\sqrt{\xi_t})^2,
  & H_n \;\nq&:=&\nq\; \sum_{t=1}^n\E[h_t]
  \\[1ex]
\label{dtmuxi}\label{DNmuxi}
  d_t(x_{<t}) \;\nq&:=&\nq\; \textstyle
  \sum_{x_t}\mu_t\ln{\mu_t \over \xi_t},
  & D_n \;\nq&:=&\nq\; \sum_{t=1}^n \E[d_t]
  \\[1ex]
\label{btmuxi}\label{BNmuxi}
  b_t(x_{<t}) \;\nq&:=&\nq\; \textstyle
  \sum_{x_t}\mu_t|\ln{\mu_t \over \xi_t}|,
  &B_n \;\nq&:=&\nq\; \sum_{t=1}^n \E[b_t]
\end{array}
\eeq

\section{Convergence of $\xi$ to $\mu$}\label{secConv}

For $D_n$ the following representation and bound is well known and
crucial \cite{Solomonoff:78,Li:97,Hutter:01alpha}
\beq\label{entropy}
  D_n \;\equiv\; \sum_{t=1}^n
  \E[d_t(x_{<t})] \;=\;
  \E[\ln{\mu(x_{1:n}) \over \xi(x_{1:n})}] \;\leq\;
  \ln{w_\mu^{-1}} \;<\; \infty
\eeq

\noindent The inequality follows from (\ref{unixi}).
The following theorem summarizes various bounds and
convergence results needed later. The major new part is Theorem
\ref{thConv}$(iv)$ which allows for an elementary proof of
$\xi_t/\mu_t\to 1$ w.p.1 based on the
Hellinger distance.

\ftheorem{thConv}{Convergence of $\xi$ to $\mu$}{
Let there be sequences $x_1x_2...$ over a finite alphabet $\X$
drawn with probability $\mu(x_{1:n})$ for the first $n$ symbols.
The mixture conditional probability $\xi'_t:=\xi(x'_t|x_{<t})$
of the next symbol $x'_t$ given $x_{<t}$ 
is related to the true conditional probability
$\mu'_t:=\xi(x'_t|x_{<t})$ in the following way:
\beqn
\begin{array}{rl}
   i) & \sum_{t=1}^n\E[\sum_{x'_t}
        (\mu'_t-\xi'_t)^2]
         \equiv  S_n  \leq
        D_n  \leq  \ln{w_\mu^{-1}}  <  \infty \\[1ex]
  ii) & \sum_{x'_t}
        (\mu'_t-\xi'_t)^2
         \equiv  s_t(x_{<t})  \leq
        d_t(x_{<t}) \toinfty{t} 0 \quad\mbox{w.p.1}\quad
        \\[1ex]
 iii) & \xi'_t - \mu'_t \to 0
        \quad\mbox{for $t\to\infty$ w.p.1 (and i.m.s) for any $x'_t$}\quad
        \\[0ex]
  iv) & \sum_{t=1}^n\E[(\sqrt{{\xi_t\over\mu_t}}-1)^2]  \leq
        H_n  \leq
        D_n  \leq  \ln{w_\mu^{-1}}  <  \infty \\[0ex]
   v) & \sqrt{{\xi_t\over\mu_t}} \to 1
        \quad\mbox{i.m.s}\qmbox{and}
        {\xi_t\over\mu_t} \to 1
        \quad\mbox{w.p.1}\quad\mbox{for $t\to\infty$} \\[1ex]
  vi) & b_t\!-\!d_t  \leq  a_t  \leq  \sqrt{2d_t},\quad
        B_n\!-\!D_n  \leq  A_n  \leq  \sqrt{2nD_n},
\end{array}
\eeqn
where $\mu_t$, $\xi_t$ are defined in (\ref{defmuxit}),
$d_t$, $D_n$ are the relative entropies (\ref{dtmuxi}), and
$w_\mu$ is the weight (\ref{xidefsp}) of $\mu$ in $\xi$.
}

\subsection{Proof} The inequality in $(ii)$ follows from the definitions
(\ref{distdD}) and from the entropy inequality
$s\leq d$ (\ref{lemEIneq}). From the definition and finiteness of
$D_\infty$ (\ref{entropy}) and from $d_t(x_{<t}) \geq 0$ one sees
that $\sqrt{d_t(x_{<t})}\stackrel{i.m.s.}\longrightarrow 0$ for $t
\to \infty$, which implies
$d_t(x_{<t})\stackrel{w.p.1}\longrightarrow 0$.
The (first) inequality in $(i)$ follows from $(ii)$ by taking the
$\E$ expectation and the $\sum_{t=1}^n$ sum. $(iii)$ follows from
$(i)$ by dropping $\sum_{x'_t}$. $(iv)$ and $(v)$ are related to
$(i)$ and $(iii)$, but are incomparable convergence results.
$(iv)$ is proven as follows:
\bqa\label{qconv}\label{eiabs}\textstyle
  \E_t[(\sqrt{\xi_t\over \mu_t}-1)^2] =
  \sum_{x_t}\nq'\;\;\mu_t
  (\sqrt{\xi_t\over \mu_t}-1)^2 =
  \\\nonumber\textstyle =
  \sum_{x_t}\nq'\;\;(\sqrt{\xi_t}-\sqrt{\mu_t})^2 \leq
  h_t(x_{<t})\leq
  d_t(x_{<t}).
\eqa
The inequalities follow from (\ref{htmuxi}) and $h\leq d$
(\ref{lemEIneq}). $(iv)$ now follows by taking the $\E$
expectation and the $\sum_{t=1}^n$ sum. $(v)$ follows from $(iv)$
by the definition of convergence i.m.s., which implies convergence
w.p.1.
The first two inequalities in $(vi)$ immediately follow from
inequalities (\ref{lemEIneq}) and definitions (\ref{distdD}).
The third inequality of $(vi)$ follows from the first by linearity
of $\E$ and $\sum$. The last inequality follows from
\bqa\label{AleqsqD}\textstyle
  \odn A_n \equiv {1\over n}\sum_{t=1}^n\E[a_t] \leq
  {1\over n}\sum_{t=1}^n\E[\sqrt{2d_t}] \leq
  \\ \nonumber\textstyle \leq
  {1\over n}\sum_{t=1}^n\sqrt{\E[2d_t]} \leq
  \sqrt{{1\over n}\sum_{t=1}^n\E[2d_t]} \equiv
  \sqrt{{\textstyle{2\over n}}D_n}
\eqa
where we have used Jensen's inequality for exchanging the averages
($\odn\sum_{t=1}^n$ and $\E$) with the concave function
$\sqrt{\;\;}$. \qed

Since the conditional probabilities are the basis of the
prediction algorithms considered in the next section and $\xi'_t$
converges rapidly to $\mu'_t$, we expect a good prediction
performance if we use $\xi$ as a guess of $\mu$. Performance
measures are defined in the next section.

Without the use of the Hellinger distance, a somewhat weaker
statement than $(v)$ can be derived from $(vi)$:
\beqn\textstyle
  \E|\ln{\mu_t\over \xi_t}| =
  \E[b_t] \leq
  \E[d_t]\!+\!\E[\sqrt{2d_t}] \leq
  \E[d_t]\!+\!\sqrt{2\E[d_t]} \toinfty{t} 0,
\eeqn
since $\E[d_t]\to 0$. I.e.\ $\sqrt{|\ln{\mu_t\over
\xi_t}|}\stackrel{i.m.}\longrightarrow 0$, which implies
${\xi_t\over\mu_t}\stackrel{i.p.}\longrightarrow 1$. The explicit
appearance of $n$ in the last expression of $(vi)$ prevents
proving stronger convergence of $\xi_t/\mu_t$ w.p.1 from $(vi)$.
Similarly \cite[Th.2]{Barron:00} shows (in our notation)
convergence of $\ln{\mu(x_{1:t})\over\xi(x_{1:t})}$ in $L_1$-norm,
which implies ${\xi_t\over\mu_t}\stackrel{i.p.}\longrightarrow
1$, but is also not strong enough to derive $(v)$.

The elementary proof for $(v)$ w.p.1 given here does not rely on
the semi-martingale convergence Theorem \cite[pp.
324--325]{Doob:53} as the proof of G\'acs in
\cite[Th.5.2.2]{Li:97}. Furthermore, $(iv)$ (and $(i)$) give a
``rate'' of convergence in the sense that the number of times
$\xi_t$ can depart from $\mu_t$ by more than $\eps$ in the sense
of $|\sqrt{\xi_t/\mu_t}-1|>\eps$ (or $|\xi'_t-\mu'_t|>\eps$) is
bounded by $\eps^{-2}\ln w_\mu^{-1}$. Note also the subtle
difference between $(iii)$ and $(v)$. If $x_{1:\infty}$ is a
$\mu$-random sequence, and $x'_{1:\infty}$ is {\em any} (possibly
constant and not necessarily $\mu$-random) sequence then
$\mu'_t-\xi'_t$ converges to zero, but no statement is possible
for $\xi'_t/\mu'_t$, since $\lim\,\inf\mu'_t$ could be zero. On
the other hand, if we stay {\em on} the $\mu$-random sequence
($x'_{1:\infty} = x_{1:\infty}$), $(v)$ shows that $\xi_t/\mu_t
\to 1$ (whether $\inf\mu_t$ tends to zero or not does not matter).
Indeed, it is easy to see that $\xi(1|0_{<t})/\mu(1|0_{<t})\propto
t\to\infty$ diverges for $\M=\{\mu,\nu\}$, $\mu(1|x_{<t}):=\odt
t^{-3}$ and $\nu(1|x_{<t}):=\odt t^{-2}$, although $0_{1:\infty}$ is
$\mu$-random \cite{Hutter:01op}.

An interesting open question is whether $\xi$ converges to $\mu$
(in difference $(iii)$ or ratio $(v)$) individually for all
Martin-L\"{o}f (M.L.) random sequences. Convergence M.L. implies
convergence $w.p.1$, but the converse may fail on a set of
sequences with $\mu$-measure zero. A convergence M.L.\ result
would be particularly interesting for $\M$ being the set of all
enumerable semimeasures and $\xi$ being Solomonoff's universal
prior. Vovk's interesting results \cite{Vovk:87} are not strong
enough to settle this point, and the proof given in
\cite{Vitanyi:00} is incomplete. See \cite{Hutter:01op} for
further discussions.

\section{Loss Bounds}\label{secLoss}

\subsection{Setup}
A prediction is very often the basis for some decision. The
decision results in an action, which itself leads to some reward
or loss. We assume that the action itself does not influence the
environment. Let $\ell_{x_t y_t}\in\Set{R}$ be the received loss
when acting $y_t\in\Y$, and $x_t\in\X$ is the actual outcome. In
many cases the prediction of $x_t$ can be identified or is already
the action $y_t$. $\X \equiv\Y$ in these cases. For convenience we
name an action a prediction in the following, even if $\X \neq\Y$.
The true probability of the next symbol being $x_t$, given
$x_{<t}$, is $\mu(x_t|x_{<t})$. The expected loss when
predicting $y_t$ is $\E_t[\ell_{x_t y_t}]$. The goal is to
minimize the expected loss. More generally we define the
$\Lambda_\rho$ prediction scheme
\beq\label{xlrdef}
  y_t^{\Lambda_\rho} \;:=\;
  \arg\min_{y_t\in\Y}\sum_{x_t}\rho(x_t|x_{<t})\ell_{x_t y_t}
\eeq
which minimizes the $\rho$-expected loss.$\!$%
\footnote{$\arg\min_y(\cdot)$ is defined as the $y$ which
minimizes the argument. A tie is broken arbitrarily. If $\Y$
is finite, then $y_t^{\Lambda_\rho}$ always exists. For infinite
action space $\Y$ we assume that a minimizing $y_t^{\Lambda_\rho}
\in \Y$ exists, although even this assumption may be removed.} As
the true distribution is $\mu$, the actual $\mu$-expected loss
when $\Lambda_\rho$ predicts the $t^{th}$ symbol and the total
$\mu$-expected loss in the first $n$ predictions are
\beq\label{rholoss}
  l_t^{\Lambda_\rho}(x_{<t}) \;:=\;
  \E_t[\ell_{x_t y_t^{\Lambda_\rho}}]
  ,\quad
  L_n^{\Lambda_\rho} \;:=\; \sum_{t=1}^n
  \E[l_t^{\Lambda_\rho}(x_{<t})].
\eeq
Let $\Lambda$ be {\em any} (causal) prediction scheme
(deterministic or probabilistic does not matter) with no
constraint at all, predicting {\em any} $y_t^\Lambda \in \Y$ with
losses $l_t^{\Lambda}$ and $L_n^{\Lambda}$ similarly defined as
(\ref{rholoss}). If $\mu$ is known, $\Lambda_\mu$ is obviously the
best prediction scheme in the sense of achieving minimal expected
loss
\beq\label{Lmuopt}
  L_n^{\Lambda_\mu} \;\leq\;L_n^{\Lambda} \quad\mbox{for any}\quad
  \Lambda.
\eeq
We prove the following loss bound for the $\Lambda_\xi$
predictor based on mixture $\xi$:

\ftheorem{thULoss}{Loss bound}{
Let there be sequences $x_1x_2...$ over a finite alphabet
$\X$ drawn with probability $\mu(x_{1:n})$ for the first
$n$ symbols. A system taking action (or predicting) $y_t \in \Y$ given
$x_{<t}$ receives loss $\ell_{x_t y_t} \in [0,1]$ if $x_t$ is the true
$t^{th}$ symbol of the sequence. The $\Lambda_\rho$-system
(\ref{xlrdef}) acts (or predicts)
as to minimize the $\rho$-expected loss.
$\Lambda_\xi$ is the prediction scheme
based on the mixture $\xi$. $\Lambda_\mu$ is the optimal
informed prediction scheme. The total $\mu$-expected losses
$L_n^{\Lambda_\xi}$ of $\Lambda_\xi$ and $L_n^{\Lambda_\mu}$ of
$\Lambda_\mu$ as defined in (\ref{rholoss}) are bounded in the
following way
\beqn\label{th3}
  0 \leq L_n^{\Lambda_\xi}-L_n^{\Lambda_\mu} \leq
  D_n+\sqrt{4L_n^{\Lambda_\mu}D_n+D_n^2} \leq
  2D_n+2\sqrt{L_n^{\Lambda_\mu}D_n}
\eeqn
where the relative entropy $D_n$ (\ref{entropy}) is bounded by
$\ln w_\mu^{-1}<\infty$.
}

The implications of Theorem \ref{thULoss} can best be read off from the
following corollary.

\fcorollary{coULoss}{Loss bound}{
Under the same conditions as in Theorem \ref{thULoss} the following
relations hold
\beqn
\begin{array}{rl}
    i)  & L_\infty^{\Lambda_\xi} \mbox{ is finite }
          \Longleftrightarrow\;\;
          L_\infty^{\Lambda_\mu} \mbox{ is finite,} \\[1ex]
   ii)  & L_\infty^{\Lambda_\xi} \leq 2D_\infty \leq 2\ln w_\mu^{-1}
          \mbox{ for \iftwocol det. \else deterministic \fi $\mu$ if }\;
          \forall x\exists y\ell_{xy}=0, \\[1ex]
  iii)  & L_n^{\Lambda_\xi}/L_n^{\Lambda_\mu}
          \iftwocol = 1+O((L_n^{\Lambda_\mu})^{-1/2})
          \rightarrow 1 \;\mbox{for}\; L_n^{\Lambda_\mu}\to\infty,
          \else \quad \!=\; 1+O((L_n^{\Lambda_\mu})^{-1/2})
          \longrightarrow 1 \quad\mbox{for}\quad L_n^{\Lambda_\mu}\to\infty,
          \fi
          \\[0ex]
   iv)  & L_n^{\Lambda_\xi}-L_n^{\Lambda_\mu} \iftwocol \;=\; \else =\qquad\fi O(\sqrt{L_n^{\Lambda_\mu}}),
          \\[-0.7ex]
\end{array}
\eeqn
Let $\Lambda$ be {\em any} prediction scheme.
\beqn
\begin{array}{rl}
    v)  & L_n^{\Lambda_\mu}\leq L_n^{\Lambda},
          \qquad\qquad\qquad\qquad\qquad\qquad\qquad\qquad\qquad\qquad\qquad \\[0ex]
   vi)  & L_n^{\Lambda} \;\geq\; L_n^{\Lambda_\xi} -
          2\sqrt{L_n^{\Lambda_\xi}D_n} \;\geq\;
          L_n^{\Lambda_\xi} - O(\sqrt{L_n^{\Lambda_\xi}}), \\[1ex]
  vii)  & L_n^{\Lambda_\xi}/L_n^{\Lambda} \quad\;\leq\; 1+O((L_n^{\Lambda})^{-1/2}).
          \\[1ex]
\end{array}
\eeqn
}
The Corollary is a trivial consequence of Theorem \ref{thULoss}
and (\ref{Lmuopt}). $(vi)$ follows from
Theorem \ref{thULoss} by replacing $L_n^{\Lambda_\mu}$ with
$L_n^\Lambda$ and solving the quadratic inequality w.r.t.\
$L_n^\Lambda$.
The main message is that the total loss $L_\infty^{\Lambda_\xi}$
of the mixture $\Lambda_\xi$ predictor is finite if the total
loss $L_\infty^{\Lambda_\mu}$ of the informed $\Lambda_\mu$
predictor is finite, and that
$L_n^{\Lambda_\xi}/L_n^{\Lambda_\mu}\to 1$ if
$L_\infty^{\Lambda_\mu}$ is not finite. $(vi)$ shows that {\em no}
(causal) predictor $\Lambda$ whatsoever achieves significantly
less (expected) loss than $\Lambda_\xi$. Worst case bounds for
aggregating strategies, especially the one derived in
\cite{Cesa:97}, explicitly depend on the comparison class. There
are always predictors which perform significantly better than the
aggregating strategy. On the other hand these algorithms have the
remarkable property that the bounds hold for {\em any} sequence,
whereas our bounds only hold in an expected sense and depend on
the environment $\mu\in\M$. See \cite{Hutter:01loss} for a more
detailed discussion of the bounds in general and this duality in
particular.

\subsection{Loss Bound of Merhav \& Feder}
The first general loss bound with no structural assumptions on
$\mu$ and $\ell$ (except boundedness) has been derived in a survey
paper by Merhav and Feder in \cite[Sec.3.1.2]{Feder:98}.
(The special case of the error-loss has earlier been
considered in \cite{Barron:93}). They showed that the regret
$L_n^{\Lambda_\xi}-L_n^{\Lambda_\mu}$ is bounded by
$\ell_{max}\sqrt{2nD_n}$ for $\ell\in[0,\ell_{max}]$. Assuming
$\ell_{max}=1$ (general $\ell_{max}$ can be recovered by scaling)
their bound reads (in our notation)
\beq\label{fmloss}
  L_n^{\Lambda_\xi}-L_n^{\Lambda_\mu} \;\leq\;
  A_n \;\leq\; \sqrt{2nD_n}.
\eeq
In Section \ref{secILoss} we prove
$
  l_t^{\Lambda_\xi}(x_{<t})-l_t^{\Lambda_\mu}(x_{<t}) \leq
  a_t(x_{<t}) \leq \sqrt{2d_t(x_{<t})}
$. Taking the the expectation $\E$ and the average
$\odn\sum_{t=1}^n$ and using Theorem \ref{thConv} shows
(\ref{fmloss}).

Bound (\ref{fmloss}) and our bound (Theorem \ref{thULoss}) are in
general incomparable. Since $2D_\infty$ is finite and
$L_n^{\Lambda_\mu}\leq n$, bound (\ref{fmloss}) can be at best a
factor $\sqrt{2}$ and an additive constant better than our bound.
On the other hand, for large $n$ and for $L_n^{\Lambda_\mu} <
{n\over 2}$ our bound is tighter. The latter condition is
satisfied if the best predictor $\Lambda_\mu$ suffers small
instantaneous loss $<\odt$ on average. Significant improvement
occurs if $L_n^{\Lambda_\mu}$ does not grow linearly with $n$, but
is for instance finite (see Corollary \ref{coULoss}, especially\
$(i)$ and $(ii)$).

\subsection{Example loss functions}
The case $\X \equiv\Y$ with unit error assignment $\ell_{xy} = 1 -
\delta_{xy}$ ($\delta_{xy} = 1$ for $x = y$ and $\delta_{xy} = 0$
for $x \neq y$) has already been discussed and proven in
\cite{Hutter:01alpha}. In this case $L_n^{\Lambda_\rho} \equiv
E_n^{\Theta_\rho}$ is the total expected number of prediction
errors. For $\X = \Y = \{0,1\}$, $\Lambda_\rho$ is a threshold
strategy with $y_t^{\Lambda_\rho}
 =  \arg\min_{y\in\{0,1\}}\{\rho_1 \ell_{1y}+\rho_0 \ell_{0y}\}
 = 0/1$ for $\rho_1\,_<^>\,\gamma$, where
$\gamma := {\ell_{01}-\ell_{00}\over
\ell_{01}-\ell_{00}+\ell_{10}-\ell_{11}}$ and $\rho_i = \rho(i|x_{<t})$. In the special error case
$\ell_{xy} = 1 - \delta_{xy}$, the bit with the highest $\rho$
probability is predicted ($\gamma = \odt$).
In the following we consider some standard loss functions for
binary outcome $\X = \{0,1\}$ and continuous action
$y$ in the unit interval $\Y = [0,1]$.
The {\em absolute loss} is defined as
$\ell_{xy} = |x - y| \in [0,1]$. The $\Lambda_\rho$ scheme
predicts $y_t^{\Lambda_\rho}  =
\arg\min_{y\in[0,1]}\{\rho_1(1 - y)+\rho_0 y\}  = 0/1$ for
$\rho_0\,_<^>\,\rho_1$. Since all predictions $y$ lie in the
subset $\{0,1\}\subset[0,1]$ and $|x - y| = 1 - \delta_{xy}$
for $y \in \{0,1\}$ this case coincides with the binary error
case above. The same holds for the $\alpha$-loss $|x-y|^\alpha$
with $0 < \alpha \leq 1$. The $\mu$-expected loss is
$l_t^{\Lambda_\rho} = \mu(i|x_{<t})$ for the $i$ with $\rho_i > \odt$.
For the {\em quadratic loss} $\ell_{xy} = (x - y)^2 \in [0,1]$ the
action/prediction $y_t^{\Lambda_\rho}  =
\arg\min_{y\in[0,1]}\{\rho_1(1 - y)^2+\rho_0 y^2\}  = \rho_1$
is proportional to the $\rho$-probability of $x_t = 1$ and
$l_t^{\Lambda_\rho} = \E_t(1-\rho(x_t|x_{<t}))^2$. For the $\alpha$-loss $|x-y|^\alpha$ with $\alpha > 1$ we get
$y_t^{\Lambda_\rho}=(1 + \sqrt[\alpha-1]{\scriptstyle \rho_0/\rho_1})^{-1}$.
For arbitrary finite alphabet
$\X$ and vector-valued predictions $\vec y$ the
quadratic loss may be generalized to $\ell_{x\vec y} = \odt \vec
y^T{\bf A}_x\vec y + \vec b_x^T\vec y + c_x$.
The {\em Hellinger loss} can be written for binary outcome in the
form $\ell_{xy} = 1 - \sqrt{|1 - x - y|} \in [0,1]$ with
$y_t^{\Lambda_\rho}  =  {\rho_1^2/(\rho_0^2+\rho_1^2)}$
and $l_t^{\Lambda_\rho} = 1-{(\mu_0\rho_0 + \mu_1\rho_1)/
\sqrt{\rho_0^2+\rho_1^2}}$.
The {\em logarithmic loss}
$\ell_{xy} = -\ln|1 - x - y| \in [0,\infty]$ is unbounded.
But since the corresponding action is $y_t^{\Lambda_\rho}
 = \rho_1$ the expected loss is
$l_t^{\Lambda_\rho} = - \E_t\ln
\rho(x_t|x_{<t})$. Hence
$l_t^{\Lambda_\xi} - l_t^{\Lambda_\mu}=d_t$ and the total loss
excess $L_n^{\Lambda_\xi} - L_n^{\Lambda_\mu} = D_n \leq
\ln{w_\mu^{-1}}$ is finitely bounded anyway and Theorem
\ref{thULoss} is not needed.

\section{Loss Bound Proof}\label{secLProof}

\subsection{Main steps}\label{subsecMainProof}
The first inequality in Theorem \ref{thULoss} has already been
proven (\ref{Lmuopt}). For the second and last inequality, we
start looking for constants $A >
0$ and $B > 0$, which satisfy the linear inequality
\beq\label{Eineq3}
  L_n^{\Lambda_\xi} \;\leq\; (A+1)L_n^{\Lambda_\mu} + (B+1)D_n.
\eeq
If we could show
\beq\label{eineq3}
  l_t^{\Lambda_\xi}(x_{<t}) \;\leq\;
  A'l_t^{\Lambda_\mu}(x_{<t}) + B'd_t(x_{<t})
\eeq
with $A':=A+1$ and $B':=B+1$
for all $t\leq n$ and all $x_{<t}$, (\ref{Eineq3}) would follow
immediately by summation and the definition of $L_n$ and $D_n$.
With the abbreviations the $m=y_t^{\Lambda_\mu}$ and
$s=y_t^{\Lambda_\xi}$ and the abbreviations (\ref{xydef}) and
(\ref{defmuxit}) the loss and entropy can then be expressed by
$l_t^{\Lambda_\xi}=\sum_i y_i \ell_{is}$,
$l_t^{\Lambda_\mu}=\sum_i y_i \ell_{im}$ and $d_t=\sum_i
y_i\ln{y_i\over z_i}$. Inserting this into (\ref{eineq3}) we get
\beq\label{lossineqa}
  \sum_{i=1}^N y_i \ell_{is} \;\leq\;
  A'\sum_{i=1}^N y_i \ell_{im} + B'\sum_{i=1}^N y_i\ln{y_i\over z_i}
\eeq
By definition (\ref{xlrdef}) of $y_t^{\Lambda_\mu}$ and $y_t^{\Lambda_\xi}$
we have
\beq\label{lcnstr}
 \sum_i y_i \ell_{im}\!\leq\!\sum_i y_i \ell_{ij} \quad\mbox{and}\quad
 \sum_i z_i \ell_{is}\!\leq\!\sum_i z_i \ell_{ij}
\eeq
for all $j$. Actually, we need the first constraint only for $j = s$ and
the second for $j = m$.
In the final paragraph of this section 
we reduce the problem to the binary $N = 2$ case, which we will
consider in the following. We take $\sum_{i=0}^1$ instead of
$\sum_{i=1}^2$ for convenience.
\beq\label{lossineqf2}
  B'\sum_{i=0}^1 y_i\ln{y_i\over z_i} +
  \sum_{i=0}^1 y_i(A'\ell_{im}\!-\!\ell_{is}) \;\stackrel?\geq\; 0
\eeq
The cases $\ell_{im} > \ell_{is}\forall i$ and $\ell_{is} > \ell_{im}\forall
i$ contradict the first/second inequality (\ref{lcnstr}).
Hence we can assume $\ell_{0m} \geq \ell_{0s}$ and
$\ell_{1m} \leq \ell_{1s}$. The symmetric case $\ell_{0m} \leq \ell_{0s}$ and
$\ell_{1m} \geq \ell_{1s}$ is proven analogously or can be reduced to the
first case by renumbering the indices ($0\leftrightarrow 1$).
Using the abbreviations $a := \ell_{0m} - \ell_{0s}$, $b := \ell_{1s} - \ell_{1m}$,
$c := y_1\ell_{1m} + y_0\ell_{0s}$, $y = y_1 = 1 - y_0$ and
$z = z_1 = 1 - z_0$ we can write (\ref{lossineqf2}) as
\beq\label{lossineqf3}\textstyle
  f(y,z) \;:=\;
\eeq
\beqn\textstyle
  B'[y\ln{y\over z}+(1\!-\!y)\ln{1-y\over 1-z}]
  + A'(1\!-\!y)a-yb+Ac \;\stackrel?\geq\; 0
\eeqn
for $zb \leq (1-z)a$ and $0 \leq a,b,c,y,z \leq 1$.
The constraint (\ref{lcnstr}) on $y$ has been dropped since
(\ref{lossineqf3}) will turn out to be true for all $y$.
Furthermore, we can assume that $d := A'(1-y)a-yb \leq 0$
since for $d > 0$, $f$ is trivially positive.
Multiplying $d$ with a constant $\geq 1$ will decrease $f$.
Let us first consider the case $z \leq \odt$. We multiply the $d$
term by $1/b \geq 1$, i.e. replace it with $A'(1-y){a\over b}-y$.
From the constraint on $z$ we known that ${a\over
b} \geq {z\over 1-z}$. We can decrease $f$ further by replacing
${a\over b}$ by ${z\over 1-z}$ and by dropping $Ac$.
Hence, (\ref{lossineqf3}) is proven for $z \leq \odt$ if we can prove
\beq\label{lossineq1}\textstyle
  f_1(y,z) := B'[ \iftwocol ... \else y\ln{y\over z}+(1\!-\!y)\ln{1-y\over 1-z} \fi] 
  + A'(1\!-\!y){z\over 1-z}-y \;\stackrel?\geq\; 0 \;\mbox{for}\;
  z\leq\odt.
\eeq
In the next paragraph of this section 
we prove that it holds for $B \geq\odA+1$. The case $z \geq \odt$
is treated similarly. We scale $d$ with $1/a \geq 1$, i.e. replace
it with $A'(1-y)-y{b\over a}$. From the constraint on $z$ we know
that ${b\over a} \leq {1-z\over z}$. We decrease $f$ further by
replacing ${b\over a}$ by ${1-z\over z}$ and by dropping $Ac$.
Hence (\ref{lossineqf3}) is proven for $z \geq \odt$ if we can
prove
\beq\label{lossineq2}\textstyle
  f_2(y,z) := B'[ \iftwocol ... \else y\ln{y\over z}+(1\!-\!y)\ln{1-y\over 1-z} \fi] 
  + A'(1\!-\!y)-y{1-z\over z} \;\stackrel?\geq\; 0 \;\mbox{for}\;
  z\geq\odt.
\eeq
In the second next paragraph of this section 
we prove that it holds for
$B \geq\odA+1$. So in summary we proved that (\ref{Eineq3}) holds
for $B \geq\odA+1$. Inserting $B =\odA+1$ into (\ref{Eineq3})
and minimizing the r.h.s.\ w.r.t.\ $A$ leads to the last
bound of Theorem \ref{thULoss} with $A = \sqrt{D_n/L_n^{\Lambda_\mu}}$.
Actually inequalities (\ref{lossineq1})
and (\ref{lossineq2}) also hold for $B \geq \odf A+\odA$, which, by
the same minimization argument, proves the slightly tighter second
bound in Theorem \ref{thULoss}. Unfortunately, the current proof
is very long and complex, and involves
some numerical or graphical analysis for determining intersection
properties of some higher order polynomials. This or a hopefully
simplified proof will be postponed. The cautious reader may
check the inequalities (\ref{lossineq1}) and
(\ref{lossineq2}) numerically for $B = \odf A+\odA$. \qed

\subsection{Binary loss inequality for $z\leq\odt$ (\ref{lossineq1})}\label{appLossIneq1}
We now prove $f_1(y,z)\geq 0$ for $z\leq\odt$ and suitable $A'
\equiv A + 1$ and $B' \equiv B + 1$. We do this by showing that
$f_1\geq 0$ at all extremal values and  ``at'' boundaries. $f_1\to
+\infty$ for $z\to 0$, if we choose $B'>0$. For the boundary $z =
\odt$ we lower bound the relative entropy by the sum over squares
$s\leq d$ (\ref{entroineq2})
\beqn
  f_1(y,\odt)\geq 2B'(y-\odt)^2+A'(1-y)-y\geq 0 \quad\mbox{for}\quad
  B\geq\odf A+\odA
\eeqn
as can be shown by minimizing the r.h.s.\ w.r.t.\ $y$. Furthermore
for $A \geq 4$ and $B \geq 1$ we have
$f_1(y,\odt) \geq 2(1-y)(3-2y) \geq 0$. Hence
$f_1(y,\odt) \geq 0$ for $B \geq \odA+1$, since for
$A \geq 4$ it implies $B \geq 1$ and for $A \leq 4$ it
implies $B \geq \odf A+\odA$.
The extremal condition $\partial f/\partial z = 0$
(keeping $y$ fixed) leads to
\beqn
  y \;=\; y^* \;:=\;
  z\!\cdot\!{B'(1\!-\!z)+A'\over B'(1\!-\!z)+A'z}.
\eeqn
Inserting $y^*$ into the definition of $f_1$ and, again,
replacing the relative entropy by the sum over squares
($y\ln{y\over z}+(1\!-\!y)\ln{1-y\over 1-z}\geq 2(y-z)^2$),
which is a special case of $s\leq d$ (\ref{entroineq2}),
we get
\beqn\textstyle
  f_1(y^*,z) \geq 2B'(y^*\!-\!z)^2+A'(1\!-\!y^*){z\over 1-z}-y^* =
  {z(1-z)\cdot g_1(z)\over [B'(1-z)+A'z]^2},
\eeqn
\beqn\textstyle
  g_1(z) := 2B'A'^2z(1-z)+[(A'-1)B'(1-z)-A'](B'+A'{z\over 1-z}).
\eeqn
We have reduced the problem to showing $g_1 \geq 0$.
If the bracket $[...]$ is positive, then $g_1$ is positive. If the
bracket is negative, we can decrease $g_1$ by increasing ${z\over
1-z} \leq 1$ in $(B'+A'{z\over 1-z})$ to $1$. The resulting expression is now
quadratic in $z$ with minima at the boundary values $z = 0$ and
$z = \odt$. It is therefore sufficient to check
\beqn
  g_1(0)\geq (AB-1)(A+B+2)\geq 0 \qmbox{and}
\eeqn
\beqn
  g_1(\odt)\geq \odt(AB-1)(2A+B+3)\geq 0
\eeqn
which is true for $B \geq \odA$. In summary we have proved
(\ref{lossineq1}) for $B \geq \odA + 1$ and $A > 0$. \qed

\subsection{Binary loss inequality for $z\geq\odt$ (\ref{lossineq2})}\label{appLossIneq2}
We now prove we show $f_2(y,z)\geq 0$ for $z\geq\odt$ and suitable
$A' \equiv A + 1 > 1$ and $B' \equiv B + 1 > 2$ similarly as in
the last paragraph by proving that $f_2\geq 0$ at all extremal
values and ``at'' boundaries. $f_2\to +\infty$ for $z\to 1$. The
boundary $z = \odt$ has already been checked in in the last
paragraph. The extremal condition $\partial f/\partial z = 0$
(keeping $y$ fixed) leads to
\beqn
  y \;=\; y^* \;:=\; z\!\cdot\!{B'z\over (B'+1)z-1}.
\eeqn
Inserting $y^*$ into the definition of $f_2$ and replacing the
relative entropy by the sum over squares $s\leq d$
(\ref{entroineq2}), we get
\beqn\textstyle
  f_2(y^*,z) \geq 2B'(y^*-z)^2+A'(1-y^*)-y^*{1-z\over z} =
  {z(1-z)\cdot g_2(z)\over [(B'+1)z-1]^2},
\eeqn
\beqn\textstyle
  g_2(z) \;:=\; [(A'-1)B'z-A'+2z(1-z)]
  (B'\!+\!1\!-\!{1\over z})+2(1-z)^2.
\eeqn
We have reduced the problem to showing $g_2 \geq 0$.
Since $(B' + 1 - {1\over z}) \geq 0$ it is sufficient to
show that the bracket is positive. We solve $[...] \geq 0$
w.r.t.\ $B$ and get
\beqn
  B\geq {1-2z(1-z)\over z}\!\cdot\!{1\over A}+{1-z\over z}.
\eeqn
For $B \geq \odA + 1$ this is satisfied for all
$\odt \leq z \leq 1$.
In summary we have proved
(\ref{lossineq2}) for $B \geq \odA + 1$ and $A > 0$. \qed

\subsection{General loss inequality (\ref{lossineqa})}\label{appLossIneqa}
We reduce
\beq\label{lossineqf2a}
  f(\vec y,\vec z):= B'\sum_{i=1}^N y_i\ln{y_i\over z_i} +
  A'\sum_{i=1}^N y_i \ell_{im}\!-\sum_{i=1}^N y_i \ell_{is} \;\geq\; 0
\eeq
\beq\label{liac}\textstyle
  \mbox{for}\quad \sum_{i=1}^Nz_i d_i\geq 0, \quad d_i:=\ell_{im}-\ell_{is}
\eeq
to the binary $N = 2$ case. We do this by keeping $\vec y$ fixed and
showing that $f$ as a function of $\vec z$ is positive at all extrema
in the interior of the simplex $\Delta:=\{\vec z: \sum_iz_i=1, z_i\geq 0\}$
of the domain of $\vec z$ and ``at'' all boundaries.
First, the boundaries $z_i \to 0$ are safe as $f \to \infty$ for $B' > 0$.
Variation of $f$ w.r.t.\ to $\vec z$ leads to a minimum at $\vec z = \vec y$.
If $\sum_i z_i d_i \geq 0$, we have
\beqn
  f(\vec y,\vec y) \!=\!
  \sum_i y_i(A'\ell_{im}\!-\!\ell_{is}) \geq
  \sum_i y_i(\ell_{im}\!-\!\ell_{is}) =
  \sum_i z_id_i \geq 0.
\eeqn
In the first inequality we used $A'>1$.
If $\sum_i z_i d_i< 0$, $\vec z = \vec y$ is outside the valid
domain due to the constraint (\ref{liac}) and the valid minima are
attained at the boundary $\Delta\cap P$,
$P := \{\vec z:\sum_iz_id_i=0\}$.
We implement the constraints with the help of Lagrange multipliers
and extremize
\beqn
  L(\vec y,\vec z) := f(\vec y,\vec z) +
  B'\lambda\sum z_i + B'\mu\sum z_id_i.
\eeqn
$\partial L/\partial z_i=0$ leads to
$y_i=y_i^*:=z_i(\lambda+\mu d_i)$. Summing this equation over $i$
we obtain $\lambda=1$. $\mu$ is a function of $\vec y$ for which a
formal expression might be given. If we eliminate $y_i$ in
favor of $z_i$, we get
\beqn\textstyle
  f(\vec y^*,\vec z) \;=\;
  \sum_i c_iz_i \qmbox{with}
\eeqn
\beqn
  c_i:=(1+\mu d_i)(B'\ln(1+\mu d_i)+A'\ell_{im}-\ell_{is}).
\eeqn
In principle $\mu$ is a function of $\vec y$ but we can treat
$\mu$ directly as an independent variable, since $\vec y$ has
been eliminated.

The next step is to determine the extrema of the function $f=\sum
c_iz_i$ for $\vec z\in\Delta\cap P$. For clearness we state the
line of reasoning for $N=3$. In this case $\Delta$ is a triangle.
As $f$ is linear in $\vec z$ it assumes its extrema at the
vertices of the triangle, where all $z_i=0$ except one. But we have to take into
account a further constraint $\vec z\in P$. The plane $P$
intersects triangle $\Delta$ in a finite line (for $\Delta\cap
P=\{\}$ the only boundaries are $z_i\to 0$ which have already been
treated). Again, as $f$ is linear, it assumes its extrema at the
ends of the line, i.e. at edges of the triangle $\Delta$ on which
all but two $z_i$ are zero. With a similar line of arguments
for $N>3$ we conclude that a necessary condition for a minimum
of $f$ at the boundary is that at most two $z_i$ are non-zero.
But this implies that all but two $y_i$ are zero.
If we had eliminated $\vec z$ in
favor of $\vec y$, we could not have made the analogous
conclusion because $y_i=0$ does not necessarily imply $z_i=0$.
We have effectively reduced the problem of showing
$f(\vec y^*,\vec z)\geq 0$ to the case $N=2$. We can go back one
step further and prove (\ref{lossineqf2a}) for $N=2$, which implies
$f(\vec y^*,\vec z)\geq 0$ for $N=2$. A proof of
(\ref{lossineqf2a}) for $N=2$ implies, by the arguments given
above,
that it holds for all $N$. This is what we set out to show here. \qed

The $N=2$ case has been proven in the previous paragraphs. This
completes the proof of Theorem \ref{thULoss}. \qed

\section{Instantaneous Losses}\label{secILoss}

Since $L_n^{\Lambda_\xi} - L_n^{\Lambda_\mu}$ is not finitely
bounded by Theorem \ref{thULoss} it cannot be used directly to
conclude analogously $l_t^{\Lambda_\xi}  -  l_t^{\Lambda_\mu} \to
0$. It would follow from $\xi_t\to\mu_t$ by continuity if
$l_t^{\Lambda_\xi}$ and $l_t^{\Lambda_\mu}$ were continuous
functions of $\xi_t$ and $\mu_t$. $l_t^{\Lambda_\mu}$ is a continuous
piecewise linear concave function, but $l_t^{\Lambda_\xi}$ is an,
in general, discontinuous function of $\xi_t$ (and $\mu_t$).
Fortunately it is continuous at the one necessary  point
$\xi_t=\mu_t$. This allows to bound $l_t^{\Lambda_\xi} -
l_t^{\Lambda_\mu}$ in terms of $\xi_t-\mu_t$.

\ftheorem{thILoss}{Instantaneous Loss Bound}{
Under the same conditions as in Theorem \ref{thULoss}, for
discrete $\M$ the following relations hold for the instantaneous
losses $l_t^{\Lambda_\mu}(x_{<t})$ and $l_t^{\Lambda_\xi}(x_{<t})$
at time $t$ of the informed and mixture prediction schemes
$\Lambda_\mu$ and $\Lambda_\xi$:
\beqn
\begin{array}{rl}
    i)  &
          \sum_{t=1}^n\E[
          (l_t^{\Lambda_\xi}-l_t^{\Lambda_\mu})^2] \;\leq\;
          2D_n \;\leq\; 2\ln w_\mu^{-1} \;<\; \infty \\[1ex]
   ii)  &
          0 \leq l_t^{\Lambda_\xi}-l_t^{\Lambda_\mu} \leq
          \sum_{x_t}|\xi_t-\mu_t| \leq
          \sqrt{2d_t}
          \;\tolim_{^{w.p.1}}^{t\to\infty}\; 0. \\[-0.5ex]
  iii)  &
          0 \;\leq\; l_t^{\Lambda_\xi}-l_t^{\Lambda_\mu} \;\leq\;
          2d_t+2\sqrt{l_t^{\Lambda_\mu}\,d_t}
          \;\tolim_{^{w.p.1}}^{t\to\infty}\; 0. \\
\end{array}
\eeqn
}

\subsection{Proof} $(ii)$ follows from
\bqan
  & & \textstyle l_t^{\Lambda_\xi}(x_{<t})-l_t^{\Lambda_\mu}(x_{<t}) \;\equiv\;
  \sum_i y_i\ell_{is}-\sum_i y_i\ell_{im} \;\leq\;
\\
  &\leq& \textstyle \sum_i(y_i-z_i)(\ell_{is}-\ell_{im}) \;\leq
  \sum_i|y_i-z_i|\!\cdot\!|\ell_{is}-\ell_{im}| \;\leq\;
\\
  &\leq& \textstyle \sum_i|y_i-z_i| \;\leq\;
  \sqrt{2\sum_i y_i\ln{y_i\over z_i}} \;\equiv\;
  \sqrt{2d_t(x_{<t})}
\eqan
To arrive at the first inequality we added $\sum_i z_i(\ell_{im} -
\ell_{is})$ which is positive due to (\ref{lcnstr}). $|\ell_{is} -
\ell_{im}| \leq 1$ since $\ell \in [0,1]$. The last inequality
follows from $a\leq\sqrt{2d}$ (\ref{entroaineqn}). $(i)$ follows
by inserting $(ii)$ and using (\ref{entropy}). $(iii)$ follows
from the proof of Theorem \ref{thULoss} by inserting $B = \odA + 1
=\sqrt{l_t^{\Lambda_\mu}/d_t}+1$ into (\ref{eineq3}). Convergence
to zero holds for $\mu$ random sequences, i.e.\ w.p.1, since
$l_t^{\Lambda_\mu} \leq 1$ is bounded. The losses
$l_t^{\Lambda_\rho}(x_{<t})$ itself need not to converge. \qed

Note, that the inequalities in $(ii)$ and $(iii)$ hold for all
individual sequences. The sum/average is only taken over the
current outcome $x_t$, but the history $x_{<t}$ is fixed. Bound
$(ii)$ and $(iii)$ are in general incomparable, but for large $t$
and for $l_t^{\Lambda_\mu} < \odt$ (especially if
$l_t^{\Lambda_\mu} \to 0$) bound $(iii)$ is tighter than bound
$(ii)$.

\section{Conclusions}\label{secConc}

\subsection{Generalization}
The only assumptions we made in this work were that $\mu\in\M$,
the loss $\ell$ is bounded to $[0,1]$, and that the decision $y_t$
does not influence the environment, i.e.\ $\mu$ is independent
$y_t$. No other structural assumptions on $\M$ and $\ell$ have
been made. The case $\mu\not\in\M$ is briefly discussed in
\cite{Hutter:01op} and more intensively in \cite{Gruenwald:98} in
a related context. Simple scaling allows loss functions in
arbitrary bounded interval \cite{Hutter:01loss}. Asymptotic
loss/value bounds for an acting agent influencing the environment
can be found in \cite{Hutter:02selfopt}.

\subsection{Optimality properties}
In \cite{Hutter:01op} we show that there
are $\M$ and $\mu\in\M$ and weights $w_\nu$ such that the derived
loss bounds are tight. This shows that the loss bounds cannot be
improved in general, i.e.\ without making extra assumptions on
$\ell$, $\M$, or $w_\nu$. We also show Pareto-optimality of $\xi$
in the sense that there is no other predictor which performs
better or equal in all environments $\nu\in\M$ and strictly better
in at least one. Optimal predictors (in a decision theoretic
sense) can always be based on a mixture distribution $\xi$. This
still leaves open how to choose the weights. We give an Occam's
razor argument that the choice $w_\nu\sim 2^{-K(\nu)}$,
where $K(\nu)$ is the length of the shortest program describing
$\nu$, is optimal.

\subsection{Outlook}
The presented Theorems and proofs are independent of the size of
$\X$ and hence should generalize to countably infinite and continuous
alphabets under (minor) technical conditions. An infinite
prediction space $\Y$ was no problem at all as long as we assumed
the existence of $y_t^{\Lambda_\rho} \in \Y$ (\ref{xlrdef}), but
even this is not essential.
The $\Lambda_\rho$ schemes and theorems may be generalized to
delayed sequence prediction, where the true symbol $x_t$ is given
only in cycle $t + d$.
Another direction is to investigate the learning aspect of
mixture prediction. Many prediction schemes explicitly learn and
exploit a model of the environment. Learning and exploitation are
melted together in the framework of universal Bayesian prediction.
A separation of these two aspects in the spirit of hypothesis
learning with MDL \cite{Li:00} could lead to new insights.
A unified picture of the loss
bounds obtained here and the loss bounds for predictors based
on expert advice (PEA) could also be fruitful.
Also, bounds which say that
the actual (not expected) loss suffered by $\Lambda_\xi$ divided
by the loss suffered by $\Lambda_\mu$ is with high probability
close to $1$ for sufficiently large $n$, would be interesting.
Maximum-likelihood predictors may also be studied.
See \cite{Hutter:01op} for further references and discussions on
the relation Bayes and PEA approaches and results,
classification tasks, games of chances, infinite alphabet,
continuous classes $\M$,
universal mixtures, and others.

\subsection{Summary}
We compared mixture predictions based on Bayes-mixes $\xi$ to
the infeasible informed predictor based on the unknown true
generating distribution $\mu$.
Convergence results of the mixture posterior $\xi_t$ to the
true posterior $\mu_t$ have been derived. A new
(elementary) derivation of the convergence in ratio has been
presented, including a rate of convergence.
The main focus was on a decision-theoretic setting, where each
prediction $y_t\in\X$ (or more generally action $y_t\in\Y$)
results in a loss $\ell_{x_ty_t}$ if $x_t$ is the true next symbol
of the sequence. We have shown that the $\Lambda_\xi$ predictor
suffers only slightly more loss than the $\Lambda_\mu$ predictor,
improving on various previous results.


\end{document}
